\documentclass[runningheads]{llncs}
\usepackage{graphicx}
\usepackage{multirow}
\begin{document}
\title{SurgeonAssist-Net: Towards Context-Aware Head-Mounted Display-Based Augmented Reality for Surgical Guidance}
\titlerunning{SurgeonAssist-Net: Context-Aware Surgical Guidance}
% \author{Anonymous}
\author{Mitchell Doughty\inst{1,2} \and
Karan Singh\inst{3} \and
Nilesh R. Ghugre\inst{1,2,4}}

% \index{Doughty, Mitchell}
% \index{Singh, Karan}
% \index{Ghugre, Nilesh R.}
%
\authorrunning{M. Doughty et al.}
\institute{Department of Medical Biophysics, University of Toronto, Toronto, Canada 
\email{mitchell.doughty@mail.utoronto.ca} \\ \and
Schulich Heart Program, Sunnybrook Health Sciences Centre, Toronto, Canada \and
Department of Computer Science, University of Toronto, Toronto, Canada \and
Physical Sciences Platform, Sunnybrook Research Institute, Toronto, Canada 
}
\maketitle              % typeset the header of the contribution
\begin{abstract}
We present SurgeonAssist-Net: a lightweight framework making action-and-workflow-driven virtual assistance, for a set of predefined surgical tasks, accessible to commercially available optical see-through head-mounted displays (OST-HMDs). On a widely used benchmark dataset for laparoscopic surgical workflow, our implementation competes with state-of-the-art approaches in prediction accuracy for automated task recognition, and yet requires $7.4\times$ fewer parameters, $10.2\times$ fewer floating point operations per second (FLOPS), is $7.0\times$ faster for inference on a CPU, and is capable of near real-time performance on the Microsoft HoloLens 2 OST-HMD. To achieve this, we make use of an efficient convolutional neural network (CNN) backbone to extract discriminative features from image data, and a low-parameter recurrent neural network (RNN) architecture to learn long-term temporal dependencies. To demonstrate the feasibility of our approach for inference on the HoloLens 2 we created a sample dataset that included video of several surgical tasks recorded from a user-centric point-of-view. After training, we deployed our model and cataloged its performance in an online simulated surgical scenario for the prediction of the current surgical task. The utility of our approach is explored in the discussion of several relevant clinical use-cases. Our code is publicly available at \url{https://github.com/doughtmw/surgeon-assist-net}.

\end{abstract}
\section{Introduction}
There has been significant interest in adoption of augmented reality (AR) for surgical guidance in the medical field, due to its ability to enhance task performance when effectively implemented \cite{peters2006image}. The use of a see-through head-mounted display (HMD) as the visualization medium, as opposed to a monitor, has been demonstrated to provide a further benefit to efficiency by eliminating the visual disconnect between the monitor and the surgical scene \cite{liu2010monitoring}. 

Though recent work has indicated the applicability of AR in laparoscopic and endoscopic procedures \cite{bernhardt2017status,zorzal2020laparoscopy}, neurosurgery \cite{meola2017augmented}, orthopedic surgery \cite{jud2020applicability}, and general surgery \cite{rahman2020head}, there remains the concern of overloading the user with too much additional information, distracting them from the task at hand and resulting in reduced performance over routine standardized techniques \cite{dixon2013surgeons}. 

\subsubsection{Motivation.} 
The bulk of research work into AR systems for medical image-guidance has centered around technical developments, including calibration \cite{grubert2017survey}, alignment \cite{peters2006image}, and visualization \cite{kersten2013state,hong2011three} and focused on achieving quantitative metrics like speed and accuracy \cite{cleary2010image}. Due to a lack of optimized virtual workflow and information representation, these advances do not guarantee the effective translation of guidance systems to clinical practice; this is evidenced by the absence of widely used commercial see-through HMD navigation systems \cite{eckert2019augmented}. 

To bridge the gap towards an effective OST-HMD based AR guidance system, our aim was to address these pitfalls by creating a framework capable of understanding the current action of a user and supporting them with only the critical information that is relevant to the task at hand, thus reducing the information overload and the need for manual control of displayed virtual content.

\subsubsection{Related Work.} 
Context-aware surgery involves the interpretation of the large amount of information created during a surgical procedure, with focus on recognizing/predicting key tasks \cite{twinanda2016endonet}, monitoring incidents \cite{suzuki2010intraoperative}, and highlighting adverse events. Surgical task prediction in an off-line context has been recently investigated in neurosurgery \cite{forestier2013multi}, laparoscopy \cite{navab2007action}, and cataract surgery \cite{quellec2014real} and has proposed the use of a secondary monitor to display virtual assistance. These applications have relied on various types of input features to predict the current surgical phase, such as radio-frequency identification chips attached to surgical instruments \cite{navab2007action}, instrument signals \cite{padoy2012statistical}, robot kinematic data \cite{lea2016learning}, external infrared measurement systems \cite{katic2015system}, or laparoscopic video \cite{twinanda2016endonet}. 

Recent applications to context-aware surgery have remained limited to procedures where high-performance computing resources and video data are readily accessible \cite{twinanda2016endonet,jin2017sv,jin2020multi}. If these systems are to be generalized to other surgical procedures, the display of context-aware information on a secondary monitor could introduce a potentially disorienting visual disconnect for the user. 

To address these challenges, Katić et al. propose a context-aware system, based on an OST-HMD, for intraoperative AR in dental implant surgery. In a porcine study, the authors demonstrated an improved task completion time and acceptance of their system by dental surgeons \cite{katic2015system}. However, the reliance on additional sensors, computing power and specialized markers for task prediction makes their method challenging to incorporate into a typical operating room workflow and incapable of generalization to different surgical scenarios. 

\begin{figure}[h]
  \includegraphics[width=\textwidth]{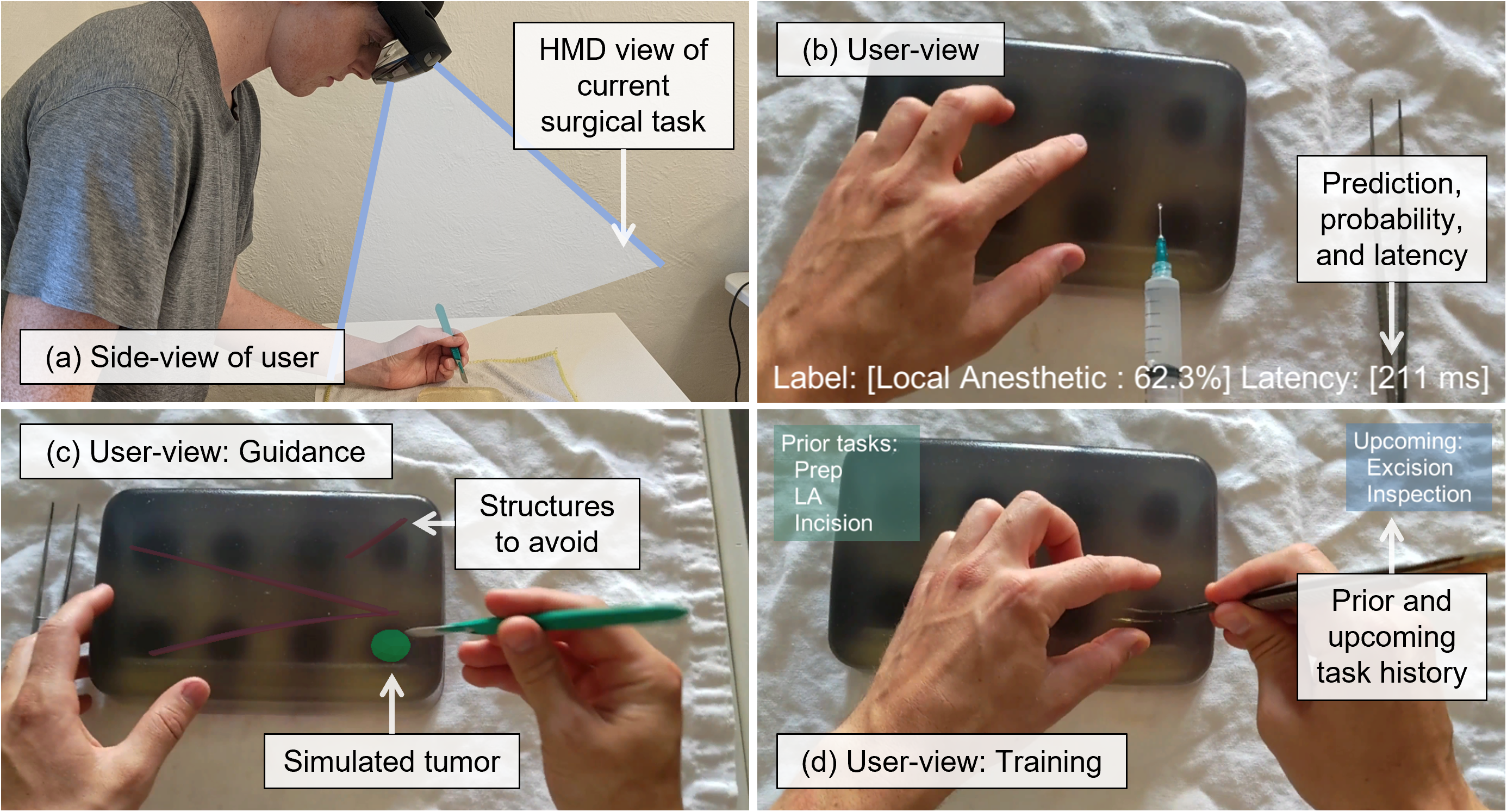}
  \caption{Example images from online inference with the SurgeonAssist-Net app on a HoloLens 2. (a) Side-view of the user and phantom. User-view with (b) information on the current surgical phase prediction, (c) minimal virtual models for task-specific guidance, and (d) information on prior and upcoming surgical tasks for user-training.} \label{sample-clinical-use-cases}
  \end{figure}

In contrast to these systems, we propose SurgeonAssist-Net, a novel framework to predict the current surgical task that a user is performing and, using the context-aware predictions, ensure that the virtual augmentation meets the current information needs of the user. Our approach does not rely on the use of external sensors, custom/specialized hardware, or additional computing power. We are the first to demonstrate the implementation of a context-aware platform on a commercially available OST-HMD with near real-time performance, eliminating the visual disconnect between a monitor and the patient (Figure \ref{sample-clinical-use-cases}). 

\section{Methods}
\subsection{SurgeonAssist-Net: Surgical Task Prediction}
\begin{figure}[t]
  \includegraphics[width=\textwidth]{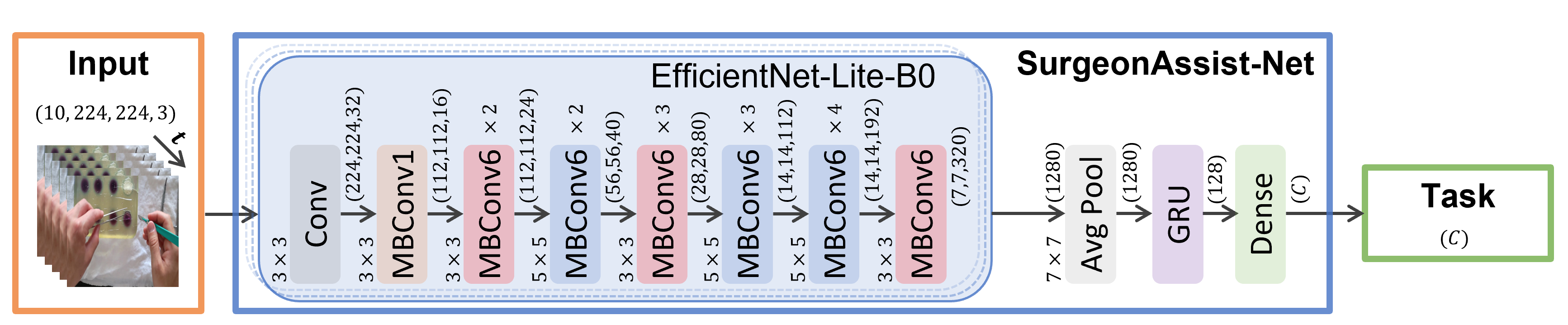}
  \caption{Overview of the deep learning-based framework for extracting relevant information from video frame data (10 frames) and predicting the current surgical task.} \label{network-layout}
  \end{figure}

To provide context-awareness to the wearer of an OST-HMD from user-centric input video, we have created the SurgeonAssist-Net framework, composed of an EfficientNet-Lite-B0 \cite{tan2019efficientnet,liu2020higher} backbone for feature extraction and a gated recurrent unit (GRU) RNN framework \cite{cho2014properties} for storing long term dependency information (Figure~\ref{network-layout}). These networks are jointly trained in an end-to-end manner, generating features that encode both spatial and temporal information. 

\subsubsection{Spatial feature extraction.} 
Deep learning and the introduction of deep CNNs has led to vast improvements in interpreting high-dimensional data over traditional approaches, finding successful applications in object detection and image recognition \cite{lecun2015deep}. With EfficientNet, Tan et al. overcame scaling issues common to increasingly deep CNNs through a compound scaling method that optimally adjusted the width, depth, and resolution of the network by using fixed coefficients, thus achieving a balance between network speed and accuracy \cite{tan2019efficientnet}. On ImageNet, EfficientNet-B0 outperforms ResNet-50 \cite{he2016deep} in top-1 and top-5 accuracy and offers a $4.9\times$ parameter and $10.5\times$ FLOPS reduction \cite{tan2019efficientnet}.

With EfficientNet-Lite-B0 \cite{liu2020higher}, modifications to the EfficientNet-B0 model were implemented to optimize performance for mobile CPU applications; we use pre-trained weights from ImageNet to serve as an initial starting point for training. The final fully connected layer at the end of the EfficientNet-Lite-B0 network was removed and replaced with a global average-pooling layer to output a $7\times7\times320$ tensor of high-level discriminative features that was reshaped to a vector of length $1280$ to serve as an input to the GRU framework (Figure \ref{network-layout}). 

\subsubsection{Temporal information modeling.} 
Recurrent neural networks can handle variable-length sequence inputs by using a hidden state augmented with non-linear mechanisms whose activation at each time step is reliant on that of the prior frame \cite{hochreiter1997long}. Both GRU and long short-term memory (LSTM) components have been used for time-series forecasting tasks like analysis of video data for activity recognition, image captioning, and surgical task prediction \cite{twinanda2016endonet}. It has been demonstrated that GRUs perform similarly to LSTM units \cite{chung2014empirical}, but have fewer total parameters and are more well-suited to real-time inference applications.

For our RNN architecture, we found optimal results using a single GRU cell with $128$ hidden units, followed by a decision network with a ReLU activation, dropout layer with probability of $0.2$, and a fully connected output layer with $C$ output nodes –- corresponding to the $C$ potential surgical tasks. The parameters of the GRU cell and dense layer were initialized using Xavier normal initialization \cite{glorot2010understanding}. During inference, we used an online recognition mode accessing only current and prior frames. After performing initial hyperparameter evaluation experiments, we found that a sequence length of $10$ video frames provided an optimal trade-off for system performance and speed (Figure \ref{network-layout}). 

\subsection{Integrating SurgeonAssist-Net for Online Inference}
We used the Microsoft HoloLens 2 OST-HMD for the recording of user-centric video and visualization of context-aware surgical task predictions. The HoloLens 2 is capable of visualization of three-dimensional (3D) virtual models through stereoscopic vision via two two-dimensional (2D) laser beam scanning displays, offering a field of view of $43\times29$ degrees (horizontal $\times$ vertical) to the wearer. 

Input frames of size $896\times504$ px were requested from the HoloLens 2 photo-video camera at a rate of $30$ frames per second (FPS), resized to $256\times256$ px using nearest-neighbor interpolation \cite{bradski2000opencv}, and center cropped to a final resolution of $224\times224$ px; these served as input to the prediction framework. We leveraged the Windows Machine Learning and Open Neural Net Exchange (ONNX) \cite{bai2019} libraries within a C\# Universal Windows Platform (UWP) app to perform inference using the SurgeonAssist-Net model. The OpenCV library \cite{bradski2000opencv} was included within a C++ UWP runtime component to prepare input frame data for prediction. The network output, the predicted task, was then used to optimize the virtual content shown to the user based on their current information needs. 

\subsection{Cholec80 Dataset}
Cholec80 contains 80 videos ($1920 \times 1080$ px or $854 \times 480$ px at $25$ FPS) of cholecystectomy surgeries performed by 13 surgeons, complete with phase annotations of the $7$ surgical phases for a procedure ($25$ FPS) defined by a senior surgeon \cite{twinanda2016endonet}. The original videos were down-sampled from $25$ FPS to $1$ FPS to match the temporal resolution used by other groups \cite{jin2020multi}. We use nearest-neighbor interpolation to scale the input video frames from the original resolution to $256\times256$ px to improve computational efficiency \cite{bradski2000opencv}. For all tests using the Cholec80 dataset, $32$ videos were used for a train split, $40$ videos for a test split, and the remaining $8$ videos for a validation split, as in prior work \cite{twinanda2016endonet}. 

Our framework was implemented using the PyTorch \cite{paszke2019pytorch} deep learning library. We trained our network for $25$ epochs with a batch size of $32$ on $2\times$ NVIDIA V100 GPUs, each with 32 GB HBM2 memory, and reported the average network performance across three training runs. For optimization, we used stochastic gradient descent (SGD), an initial learning rate of $5e^{-4}$, and a momentum of $0.9$. Sequence-wise data augmentation was performed on each training batch of image data, including random cropping of input images to $224\times224$ px, horizontal and vertical flipping, and random color augmentation. 

To evaluate the performance of the SurgeonAssist-Net for task recognition, we employed the widely used metrics of accuracy (AC), precision (PR), and recall (RE) \cite{twinanda2016endonet} and compared the results to other recent approaches.  Furthermore, we included an estimate of the total number of parameters in each model, the model size, the inference time (latency), and the FLOPS for an input image sequence of size ($t \times 224 \times 224 \times 3$), where $t$ is the input sequence length of that specific approach. A single core of an AMD Ryzen 3600 CPU was used to measure the average latency for each network over $10$ runs on a subset of the testing data.

\subsection{User-Centric Surgical Tasks Dataset}
Due to the lack of available video of surgical tasks from a user-centric point of view, we created our own dataset for training the SurgeonAssist-Net framework and evaluating its online performance on the HoloLens 2 device. The dataset included a total of five surgical tasks performed by three novice users on a gelatin phantom as they worked to remove a simulated subsurface tumor. During the task, the typical suite of surgical tools: scalpel, forceps, scissors, clamp, and syringe, was made available to the users. We recorded the dataset using the photo-video camera on the HoloLens 2 OST-HMD ($1280\times720$ px at $30$ FPS). A total of $52,500$ frames were extracted from the videos at a rate of $30$ FPS. Details of the dataset including tasks and duration are included in Table~\ref{simple-surgical-task}.

\begin{table}[t]
  \caption{Details of the user-centric surgical tasks dataset.}
  \label{simple-surgical-task}
  \centering
  \begin{tabular}{p{0.175\linewidth}|p{0.15\linewidth}|p{0.15\linewidth}|p{0.15\linewidth}|p{0.15\linewidth}|p{0.15\linewidth}} \hline
  Phase & \centering Preparation & \centering Local anes. & \centering Incision & \centering Excision & \centering Inspection \tabularnewline \hline
  Duration (sec) & \centering$27.2 \pm 11.5$ & \centering$39.2 \pm 10.0$ &\centering $15.1 \pm 9.9$ & \centering$32.3 \pm 23.5$ & \centering$12.3 \pm 8.9$ \tabularnewline
  Annotations & \centering$4,110$ & \centering$5,880$ & \centering$13,800$ & \centering$21,360$ & \centering$7,350$ \tabularnewline \hline
  \end{tabular}
\end{table}

As with the Cholec80 benchmark dataset, we resized input video frames of the user-centric surgical tasks dataset to a resolution of $256 \times 256$ px \cite{bradski2000opencv} and used an input sequence length of $10$ frames. We segmented the dataset such that 3 videos were used for a train split, 1 video for a test split, and the remaining 1 video for a validation split. 

\section{Results and Discussion}
\subsection{Cholec80: Surgical Task Prediction on a Benchmark Dataset}
Table \ref{cholec80-results-sota} compares the performance and latency of each approach on the Cholec80 dataset. Twinanda et al. have presented surgical task prediction results using learned visual features and temporal dependencies \cite{twinanda2016endonet} based on (1) the single-task PhaseNet with features extracted from a modified AlexNet backbone \cite{krizhevsky2012imagenet}; and (2) the multi-task EndoNet framework which makes use of a modified AlexNet backbone for feature extraction and tool classification; both approaches use a single image frame as the input sequence. The single-task SV-RCNet \cite{jin2017sv} and multi-task MTRCNet-CL \cite{jin2020multi} networks share a similar ResNet-50 architecture for feature extraction and an LSTM cell with $512$ hidden nodes for phase prediction. Additionally, both the SV-RCNet and MTRCNet-CL approaches use an input sequence length of $10$ frames for prediction. As we were only interested in single-task performance, we did not report the results of multi-task approaches like EndoNet and MTRCNet-CL in our evaluation.  

\begin{figure}[t]
  \begin{center}
    \includegraphics[width=\textwidth]{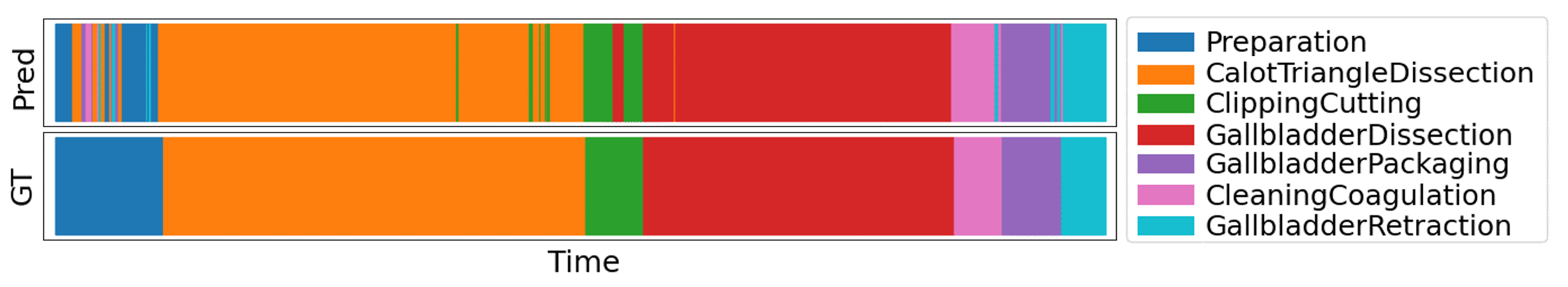}
  \end{center}
  \caption{SurgeonAssist-Net phase prediction (Pred) performance visualized relative to the ground truth (GT) phase labels on Video41 of the Cholec80 dataset (51 m 43 s in duration). The legend indicates the color coding of each individual phase.}
  \label{cholec80-vis}
\end{figure}

SurgeonAssist-Net outperformed the PhaseNet \cite{twinanda2016endonet} framework across AC, PR, and RE metrics, required $46\times$ fewer parameters for operation, and used $45\times$ less memory for model deployment. When compared to SV-RCNet \cite{jin2017sv}, SurgeonAssist-Net scored better in AC and PR metrics and required $7.4\times$ fewer model parameters, achieved $10.2\times$ faster FLOPS, and used $3\times$ less time for inference. Due to its performance efficiency, low parameter count, and compact model size, SurgeonAssist-Net can be effectively operated in computationally restricted environments for real-time inference. Figure~\ref{cholec80-vis} provides a qualitative representation of the performance of SurgeonAssist-Net, without any form of post-processing, across a subset of the Cholec80 dataset (Video41). 

\begin{table}[t]
  \caption{Results versus state-of-the-art using the Cholec80 surgical tasks dataset.}
  \label{cholec80-results-sota}
  \centering
  \begin{tabular}{p{0.175\linewidth}|p{0.09\linewidth}|p{0.075\linewidth}|p{0.075\linewidth}|p{0.075\linewidth}|p{0.075\linewidth}|p{0.1\linewidth}|p{0.1\linewidth}|p{0.1\linewidth}} \hline
  \multirow{2}{*}{Method} &	\centering Frames ($t$) &	\centering AC (\%) &	\centering PR (\%) &  \centering RE (\%) &	\centering Size (MB) &	\centering Params (M) &	\centering FLOPS (B) &	Latency (ms) \centering \tabularnewline \hline 
  Ours 				& 	\centering$10$	&	\centering$\mathbf{85.8}$ &	\centering$\mathbf{81.5}$ &		\centering$81.4$  & 	\centering$\mathbf{15.9}$ & 	\centering$\mathbf{3.91}$ &	\centering$4.04$ &	\centering$532.4$ \tabularnewline
  SV-RCNet \cite{jin2017sv}		&	\centering$10$	&	\centering$85.3$ &	\centering$80.7$ &		\centering$\mathbf{83.5}$ & 	\centering$115.3$ &	\centering$28.76$ &	\centering$41.25$ &	\centering$1593.8$ \tabularnewline
  PhaseNet \cite{twinanda2016endonet}&\centering$1$	&		\centering$73.0$&		\centering$67.0$&			\centering$63.4$  & \centering$718.8$ &		\centering$179.71$&	\centering$\mathbf{0.83}$&		\centering$\mathbf{99.3}$ \tabularnewline 
  \hline
  \end{tabular}
  \end{table}

\subsection{User-Centric Surgical Tasks Dataset: Task Prediction and Online Performance on the HoloLens 2}
To make the SurgeonAssist-Net model available to the HoloLens 2 through the UWP interface, we converted our trained model from its PyTorch implementation to ONNX format \cite{bai2019}. Table \ref{user-centric-results-sota} includes the relative performance of the SurgeonAssist-Net framework in PyTorch and ONNX formats compared with ONNX converted implementations of PhaseNet \cite{twinanda2016endonet} and SV-RCNet \cite{jin2017sv} when evaluated on the user-centric surgical tasks dataset.  

A small decrease in AC, PR, RE and model size was recorded following conversion of the SurgeonAssist-Net model to ONNX format. However, we also measured a $5.2\times$ decrease in CPU inference time by the ONNX model when compared to its PyTorch implementation; this speedup was due to the compilation of an efficient graph model during ONNX model export. Similar relative performance across AC, PR, and RE by the SurgeonAssist-Net model as compared to the ONNX converted PhaseNet \cite{twinanda2016endonet} and SV-RCNet \cite{jin2017sv} was observed. 

\subsubsection{HoloLens 2 performance and feasibility.} 
To evaluate the real-world performance of the SurgeonAssist-Net model on the HoloLens 2 headset, we created a sample application that displayed the prediction, with its associated probability and latency, for the current surgical task being performed. In Figure \ref{sample-clinical-use-cases} we include a sample image from an experiment where a user was presented with the same gelatin phantom and surgical tools as in the user-centric surgical tasks dataset and tasked with removing a subsurface tumor. Across the test, there was good agreement between the predicted and user-performed tasks. 

The latency of the SurgeonAssist-Net ONNX model on the HoloLens 2 CPU, averaged across 30 seconds of online predictions, was measured to be $219.2$ ms, or roughly $5$ FPS, with a single image input sequence. To measure the feasibility of online inference with other networks on the HoloLens 2, we repeated the above experiment with ONNX converted PhaseNet \cite{twinanda2016endonet} and SV-RCNet \cite{jin2017sv} models; however, we were unable to successfully load or operate either model on the HoloLens 2 CPU due to the large model size and/or high FLOPS requirements.

\begin{table}[t]
  \caption{Results versus state-of-the-art using the user-centric surgical tasks dataset.}
  \label{user-centric-results-sota}
  \centering
  \begin{tabular}{p{0.265\linewidth}|p{0.13\linewidth}|p{0.09\linewidth}|p{0.09\linewidth}|p{0.09\linewidth}|p{0.12\linewidth}|p{0.155\linewidth}} \hline
  Method &	\centering Frames ($t$) &	\centering AC (\%) &	\centering PR (\%) &  \centering RE (\%) &	\centering Size (MB) &	Latency (ms) \centering \tabularnewline \hline 
  Ours PyTorch 				& 	\centering$10$	&	\centering$\mathbf{85.5}$ &	\centering$\mathbf{88.4}$ &		\centering$76.5$  & 	\centering$15.9$ & \centering$538.6$ \tabularnewline
  Ours ONNX 				& 	\centering$10$	&	\centering$85.1$ &	\centering$87.5$ &		\centering$75.3$  & 	\centering$\mathbf{15.2}$ &	\centering$103.1$ \tabularnewline
  SV-RCNet ONNX \cite{jin2017sv}		&	\centering$10$	&	\centering$84.9$ &	\centering$87.3$ &		\centering$\mathbf{77.5}$ & 	\centering$112.2$ &	\centering$721.2$ \tabularnewline
  PhaseNet ONNX \cite{twinanda2016endonet}&\centering$1$	&		\centering$70.6$&		\centering$69.7$&			\centering$60.1$  & \centering$702.0$ &		\centering$\mathbf{64.7}$ \tabularnewline 
  \hline
  \end{tabular}
\end{table}

\subsection{Clinical Significance}
Aside from accuracy, a primary limitation of AR-guided approaches is the reliance upon a user to manually control the appearance and presentation of virtually augmented entities, thereby adapting the visualization to their current surgical context. This is tedious and detracts from their focus on the task. Our work on surgical task prediction is thus critical and foundational in ensuring that the automated augmentation of virtual models meets the current information needs. In this work, the predicted surgical task serves as a prerequisite to displaying the optimal virtual information to the user. We will now briefly discuss three clinical scenarios involving surgical guidance, user-training, and performance evaluation, where the SurgeonAssist-Net could be readily incorporated.

\paragraph{Guidance.} The predicted task context can control the choice and presentation of the augmented virtual models. For example, in general surgery, as a surgeon picks up a scalpel and the incision phase of a procedure is detected, the HMD would display a relevant virtual model indicating the target site for surgical entry (Figure \ref{sample-clinical-use-cases}). Throughout the procedure, our surgical phase detection would enable different virtual models and information relevant to the surgical phase to be optimally selected and presented, without user intervention. 

\paragraph{Training.} Task prediction can be used to guide a student, wearing the HMD, in practicing a general surgery task on a cadaver. Their active learning can be reinforced by presenting them with a task history of phases performed, or of upcoming surgical actions, given their present surgical step (Figure \ref{sample-clinical-use-cases}). Additional relevant information, in the form of visual cues, text, or audio, could be presented in tandem with the detected task to enhance the training experience. 

\paragraph{Evaluation.} Task analytics can provide surgeons with quantitative data on a surgical procedure. For example, a surgeon performing a less frequent procedure could wear the HMD while re-training and be provided with chronology and analytics of the time spent in each surgical phase (Figure \ref{cholec80-vis}). This information, when compared to peers, could serve to suggest focus areas for improvement.

\section{Conclusions and Future Work}
The focus of this work was to create a lightweight framework capable of understanding user-centric activities and providing virtual real-time workflow assistance for a pre-defined set of surgical tasks. By training the SurgeonAssist-Net framework on the user-centric surgical tasks dataset, we were able to use it effectively as an event-detection heuristic to associate known events in an online scenario by using the on-board computing resources of an OST-HMD (in this case, the HoloLens 2). For future investigation, we envision that an in-depth user-study to evaluate a manually-controlled AR experience versus the context-aware approach could further demonstrate the benefits of action-driven virtual assistance. We also expect that a larger user-centric training dataset would result in better consistency in predictions from the SurgeonAssist-Net framework. Nonetheless, we have demonstrated the potential capabilities of an online context-aware surgical guidance platform and brought attention to its capacity to overcome issues which had previously plagued AR-based image-guidance systems. 

\subsubsection{Acknowledgements.} This work was supported by the Natural Sciences and Engineering Research Council of Canada (NSERC) Discovery program (RGPIN-2019-06367).

\bibliographystyle{splncs04}
\bibliography{paper1825}
\end{document}